# A Method for Characterizing Disease Progression from Acute Kidney Injury to Chronic Kidney Disease

Yilu Fang, MA[1] *, Jordan G. Nestor, MD, MS[2] *, Casey N. Ta, PhD[1],

Jerard Z. Kneifati-Hayek, MD, MS[3], Chunhua Weng, PhD[1]

**Affiliations**

[1] Department of Biomedical Informatics, Columbia University, New York, NY, USA

[2] Department of Medicine, Division of Nephrology, Columbia University, New York, NY, USA

[3] Department of Medicine, Division of General Medicine, Columbia University, New York, NY, USA

* Equal contribution first authors

**Corresponding Author:**

Chunhua Weng, PhD

Department of Biomedical Informatics, Columbia University

622 W 168th St, PH20

New York, NY 10032, USA

cw2384@cumc.columbia.edu

**Keywords**:

Acute kidney injury, Chronic kidney disease, Natural language processing, Multi-state modeling, Survival analysis

## Abstract

### Objective

Patients with acute kidney injury (AKI) are at high risk of developing chronic kidney disease (CKD), but identifying those at greatest risk remains challenging. We used electronic health record (EHR) data to dynamically track AKI patients' clinical evolution and characterize AKI-to-CKD progression.

### Methods

Post-AKI clinical states were identified by clustering patient vectors derived from longitudinal medical codes and creatinine measurements. Transition probabilities between states and progression to CKD were estimated using multi-state modeling. After identifying common post-AKI trajectories, CKD risk factors in AKI subpopulations were identified through survival analysis.

### Results

Of 20,699 patients with AKI at admission, 3,491 (17%) developed CKD. We identified fifteen distinct post-AKI states, each with different probabilities of CKD development. Most patients (75%, n=15,607) remained in a single state or made only one transition during the study period. Both established (e.g., AKI severity, diabetes, hypertension, heart failure, liver disease) and novel CKD risk factors, with their impact varying across these clinical states.

### Conclusion

This study demonstrates a data-driven approach for identifying high-risk AKI patients, supporting the development of decision-support tools for early CKD detection and intervention.

## 1. INTRODUCTION

Clinicians continually seek new knowledge and a deeper understanding of how diseases present and why they progress. Researchers are called to develop methods that effectively extract clinically meaningful information from large datasets to support decision-making and personalized care.[1,2] Analyzing longitudinal real-world electronic health record (EHR) data offers an unprecedented opportunity to track disease progression and identify high-risk patients by providing a dynamic view of disease evolution and uncovering patterns that may be missed in traditional observational studies. Developing novel analytic methods is essential for improving diagnosis, risk stratification, and treatment decisions, as well as advancing precision medicine. Here, we apply a data-driven approach to characterize the transition from acute kidney injury (AKI) to chronic kidney disease (CKD), demonstrating its broader potential to extract meaningful clinical insights across diseases and patient populations.

AKI is a common and serious condition, affecting up to 50% of ICU patients and 20% of adult hospital admissions.[3-7] It contributes to 1.7 million deaths annually and increases the risk of CKD, which affects 14% of adults in the U.S. and is the leading cause of morbidity and mortality.[8] AKI survivors face a high risk of CKD progression, with incidence rates reaching 25.8 cases per 100 patient-years.[6,9-17] The high variability in AKI presentations complicates the implementation of effective prevention and intervention strategies.[7,18-20] Identifying AKI subpopulations at highest risk for CKD is critical for early intervention and personalized management.

Despite this, most studies on AKI progression rely on static, single-timepoint features such as demographics and comorbidities.[21] These approaches fail to capture dynamic, time-varying factors that influence AKI outcomes. Additionally, prior studies have largely focused on specific AKI subtypes, such as sepsis-associated AKI or AKI in COVID-19 ICU patients, often emphasizing mortality rather than CKD progression.[22-25] Some research links AKI subtypes to major adverse kidney events, including CKD, but gaps remain in identifying which patients need closer follow-up.[26]

This study addresses these limitations by leveraging longitudinal EHR data to track patient trajectories after AKI. Multi-state models, which capture transitions between distinct clinical states over time[27], have been used in studies of AKI recovery, kidney transplantation, and mental health outcomes.[28-31] Here, we apply natural language processing (NLP), clustering analysis, multi-state modeling, and survival analysis to a large EHR dataset to examine AKI-to-CKD progression. This approach allows us to define distinct post-AKI clinical states and determine which subpopulations are at highest risk for CKD.

**Statement of significance**

| | |
|---|---|
| **Problem** | Patients with acute kidney injury (AKI) are at high risk of developing chronic kidney disease (CKD), but identifying those at greatest risk remains challenging. |
| **What is Already Known** | Most studies on AKI progression rely on static, single-timepoint features, which fail to capture dynamic, time varying factors that influence AKI outcomes. |
| **What this Paper Adds** | We present a novel scalable and data-driven method for dynamically tracking AKI patients' clinical evolution and characterizing AKI-to-CKD progression using longitudinal EHR data. It identifies high-risk AKI patients and supports decision support tools for early CKD detection and intervention. |
| **Who would benefit from the new knowledge in this paper** | Biomedical informaticians who would like to characterize the progression from an acute condition to a chronic condition. |

## 2. METHODS

In this study, we 1) used NLP and clustering to define clinical states based on medical codes and creatinine trends, 2) applied multi-state modeling to quantify transition probabilities, and 3) identified common post-AKI trajectories and recognized CKD risk factors using survival analysis. Figure 1 provides an overview of the AKI-to-CKD progression characterization.

### 2.1. Data Source

This retrospective cohort study utilized the NewYork-Presbyterian/Columbia University Irving Medical Center (NYP/CUIMC) database, comprising EHR data from 7,062,954 patients spanning October 1985 to September 2023.

### 2.2. Cohort Construction

We defined an "AKI visit" as a patient's first AKI-associated hospitalization and included those with at least one documented AKI-related medical concept code (see Supplementary Table S1). Visits without creatinine values were excluded.

Numerous classification criteria for AKI were considered (e.g., RIFLE, KDIGO).[32-35] Ultimately, we defined AKI in the following way: 1. Baseline creatinine for each potentially eligible patient was determined using the following approach, as described by Xu et al.[33] and Stevens et al.[35]:

1) Median creatinine from 365 to 7 days before the visit start date
2) Minimum creatinine from 7 to 0 days before the visit start date
3) Minimum creatinine during the visit

2. Patients were considered eligible for this study if creatinine levels rose by 50% or more from baseline[35,36].

We excluded patients with concurrent CKD by a) CKD-related codes (see Supplementary Table S2) documented before or during the visit, or b) those with a baseline creatinine of ≥1.2 mg/dL at the time of the visit, indicating potential underlying CKD. Additionally, we excluded patients whose creatinine level did not rise by 50% or more above baseline within 48 hours of admission, or who had fewer than two creatinine measurements during this initial 48-hour period.

**2.3. Patient Representation**

For each patient $i$, we defined the observation time window $T_i = \left[t_i^{AKI}, t_i^{end}\right]$, where $t_i^{AKI}$ denotes the time of the first documented AKI event, and $t_i^{end}$ corresponds to the earliest of: a) a diagnosis of CKD or b) the date of last recorded entry in the database for patients with right-censored data or those affected by the competing risk of all-cause mortality. Patients who developed CKD after AKI were identified by the presence of CKD-related concept codes (see Supplementary Table S2).

Let $\Delta t_i^j = t_i^j - t_i^{AKI}$ denotes the number of days since the AKI event at the $j$-th recorded time point, for $j = 1, 2, \ldots, N_i$, where $N_i$ is the total number of days with available clinical records within the interval $T_i$ for patient $i$. Within the window $T_i$, we systematically constructed a time-indexed sequence of vectors $\mathcal{X}_i = \left\{x_i^j\right\}_{j=1}^{N_i}$, representing the cumulative patient trajectory. At each $\Delta t_i^j$, we constructed a vector $x_i^j \in \mathbb{R}^d$, representing the patient's aggregated clinical status from the day of AKI event up to and including day $\Delta t_i^j$. Each vector $x_i^j$ was formed by concatenating two embeddings: medical coding sequence embeddings $e_i^1(\Delta t_i^j) \in \mathbb{R}^{d_1}$ and temporal creatinine embeddings $e_i^2(\Delta t_i^j) \in \mathbb{R}^{d_2}$. Thus, each vector is of the same size, allowing for consistent representation in downstream analysis. It is represented as

$$x_i^j = \text{concat}\left(e_i^1(\Delta t_i^j), e_i^2(\Delta t_i^j)\right).$$

The temporal medical coding sequence embeddings $e_i^1$ capture cumulative exposure to structured clinical data, including all conditions, medications, procedures, and measurements over the interval, representing the longitudinal clinical patterns[37]. These embeddings were generated using Paragraph Vector Models trained on EHRs, as detailed in our prior publication[37]. Each embedding is of fixed length, ensuring consistent vector representations across patients regardless of sequence length.

Building upon this foundation, the present study extends the approach by constructing temporal creatinine embeddings $e_i^2$ to capture the kidney-specific physiological trajectories, providing a comprehensive representation of disease progression. We constructed a bivariate time series for patients' creatinine values and corresponding ages over the observation window. The Cr values recorded on the same date were averaged. These time series were used as input to a transformer-based framework for multivariate time series representation learning, as proposed by Zerveas et al.[38]. The model underwent pre-training through masked language modeling objective on a separate population of patients (excluded from the cohort) who had at least 50 Cr values recorded. This pre-training phase allowed the model to learn temporal dependencies and clinical patterns in longitudinal kidney function. A proportion of 0.15 of each variable sequence was masked, and the mean squared error loss of predictions made on the masked values was minimized. The model was pre-trained for 100 epochs, with the epoch exhibiting the best performance being saved. The model was then fine-tuned on the study cohort to perform a CKD diagnosis classification task. Fine-tuning was conducted for 20 epochs, and the model checkpoint corresponding to the lowest weighted binary cross-entropy loss on the validation set was selected. The weight for a class was calculated as the total number of samples in the training set divided by two and the number of samples in that class. The train-validation-test ratio for both pretraining and fine-tuning was 0.64:0.16:0.20. We compared the model's performance across various parameter settings, and the optimal configuration is outlined in Supplementary Table S3. For each patient and each time point $t_i^j$, the time series segment up to that day was input into the model, and the corresponding embedding $e_i^2(\Delta t_i^j)$ was extracted from the model's penultimate layer at each time step $\Delta t_i^j$.

## 2.4. Clinical State Identification and Characterization

To uncover distinct patterns of disease progression, we apply KMeans++ clustering over the set of all vectors across the full patient cohort and time horizon $\mathcal{X} = \{x_i^j | \forall i, \forall j \in [1, N_i]\}$. Each resulting cluster represents a clinical state. The optimal number of clusters $K$ was determined using the elbow method with the Within-Cluster Sum of Squares (WCSS) as the key metric to achieve distinct subgroups and avoid overfitting. The elbow is detected using kneed[39]. We evaluate $K$ over a range from 3 to 40. In addition to this analytical approach, we conducted a clinical investigation in collaboration with a board-certified nephrologist (JGN). Cluster results near the elbow point were reviewed for clinical plausibility and interpretability based on the clinical characteristics of each cluster. Expert feedback was incorporated into the final selection of $K$ to ensure that the resulting states aligned with meaningful patterns of disease progression. Each vector $x_i^j$ is assigned to a state, indicating the clinical state occupied by patient $i$ at cumulative day $\Delta t_i^j$.

To enhance interpretability and ensure clinical plausibility, each derived state was characterized by conditions with a prevalence >20% in at least one state (i.e., more than 20% of the vectors assigned to a particular state represent a specific time interval during which the condition codes were recorded). These conditions were further curated and validated by the board-certified nephrologist and subsequently grouped into organ-system categories, thereby providing a human-generated explanation for the machine-derived states. Ontology expansion technique[40] was applied to condition concepts before calculating their prevalence, where all ancestors of a concept in SNOMED-CT[41] were added.

Risk factors for CKD development were identified for each prevalent initial clinical state (states at cohort entry) using Cox proportional hazards models. Based on prior literature and clinical expert review, we included the following variables: sex, age at encounter, baseline estimated glomerular filtration rate (eGFR)[42,43], factors predisposing patients vulnerable to AKI documented in the literature[44-52], and initial illness severity assessment during the AKI visit. Baseline eGFR was calculated using the 2021 CKD-EPI Equation[42] for adults (age ≥ 18 years) and the equation from Pierce et al.[43] for children (age < 18).

Factors predisposing patients to AKI documented in the literature include pre-existing comorbidities such as diabetes mellitus (DM), hypertension (HTN), coronary artery disease and/or myocardial infarction (CAD_MI), congestive heart failure (CHF), peripheral vascular disease (PV), and liver disease (LD), as well as medications taken within one year of the AKI visit. These medications encompass loop and thiazide diuretics, agents affecting the renin-angiotensin system (RAS) (i.e., ACE inhibitors, angiotensin II receptor blockers (ARBs), and direct renin inhibitors), and non-steroidal anti-inflammatory drugs (NSAIDs) like ibuprofen and naproxen.[44] Their concept codes are in Supplementary Tables S4 and S5. For patients without documented diagnoses or medication use within the specific timeframe, it was assumed they did not have these conditions or take these medications.

The initial assessment of illness severity during the AKI visit includes vital signs, serum studies, and the documentation of sepsis (identified by concept codes) within the initial 24 hours of the visit. Vital signs include body temperature (Temp), systolic blood pressure (SBP), diastolic blood pressure (DBP), heart rate (HR), respiratory rate (RR), and oxygen saturation level (SpO2). Serum studies include: a) Basic metabolic panel: sodium (Na), potassium (K), bicarbonate (HCO3), blood urea nitrogen (BUN), creatinine (Cr), and glucose (Glu); b) Other electrolytes: calcium (Ca) and magnesium (Mg); c) Hepatic panel: albumin, bilirubin (total), aspartate aminotransferase (AST), alanine aminotransferase (ALT), and alkaline phosphatase (ALP); d) Complete blood count: white blood cells (WBC), hemoglobin (Hb), and platelets (Plts); e) Coagulation studies: focus on International Normalized Ratio (INR). Two experienced physicians (JGN, JKH) established biologically reasonable ranges for measurements (see Supplementary Table S6 for concept codes, unit, descriptive statistics of raw data, and established ranges). Measurements falling outside these ranges were considered invalid and excluded. Mean values were then calculated for each measurement within the initial 24 hours of the visit (i.e., average value on day 1).

Missing sex was imputed using K-nearest Neighbors with K=5[53]. All measurement variables have missingness (Supplementary Table S6). We excluded variables (i.e., SpO2, Plts, and HCO3) with a missing rate >0.5 and used SoftImpute for the remaining missingness.[54-56]

**2.5. State Transition Identification and Characterization**

Following the identification of clinical states, we analyzed transitions between clinical states to track changes in patient health status from the initial AKI visit to the earliest CKD diagnosis, considering the competing risk of all-cause mortality and cases where CKD development was not observed. We considered CKD development to have occurred if CKD medical codes were documented at any time following the AKI visit. We modeled these transitions using a time-inhomogeneous Markovian multistate model with a finite state space. In this framework, each patient trajectory is represented as a sequence of state transitions over time, where transition probabilities vary over time and future states depend only on the current state. We explicitly accounted for the competing risk of all-cause mortality, and both CKD development and death were treated as terminal states. We employed the Aalen-Johansen estimator[57] to estimate the transition probabilities between states.

We used Cox models to identify risk factors associated with transitions between clinical states and the development of CKD in different AKI subpopulations, categorized based on their first few clinical states within their transition paths. Candidate risk factors were identical to those used in the initial state analyses.

## 2.6. Statistical Analysis

We employed Kolmogorov-Smirnov tests to compare continuous variables and Chi-squared tests for categorical variables between CKD and non-CKD patients. We identified CKD risk factors using cause-specific Cox proportional hazards models, accounting for the competing risk of death. Weighted Cox proportional hazards models[58,59] were used due to violations of the proportional hazards assumption, with non-converging models excluded after 10,000 iterations. For each model tailored individually for CKD development and different initial states (or initial state transitions), we managed the multicollinearity among continuous variables by retaining only one from any set presenting a high correlation (i.e., Pearson correlation>0.7). Binary variables with less than five occurrences correlating with outcome events were excluded from the analysis. To control for multiple comparisons, we applied the Benjamini and Hochberg method for false discovery rate adjustment[60]. Sensitivity analyses excluding the competing risk of death assessed robustness. Statistical significance was set at P < 0.05 (two-sided). Python, version 3.8.13 and R software, version 4.2.1, were used for analyses.

## 2.7. Ethics Approval

This EHR-based research was determined by the CUIMC Institutional Review Board (IRB protocol number AAAR3954) to qualify for a waiver of consent as per 45CFR46.116(d) as the following criteria are met in this study:

(1) The research involves no more than minimal risk to the subjects.

The research involves analysis of existing data only, and the risk from the EHR-based research is minimal. There is a risk that participants could be harmed in the unlikely event that information was disclosed outside the study in an identifiable way. We will take a number of steps to protect the privacy and confidentiality of all our participants and to minimize the risks associated with possible distress and burden. All data analysis will be performed in secure and CUIMC-approved servers, and we will safeguard the data sets to ensure the privacy of the patients and to eliminate the risks of data leak.

(2) The waiver or alteration will not adversely affect the rights and welfare of the subjects.

The waiver will not adversely affect the rights and welfare of the study participants, because the study involves analysis of existing data without collecting new data, and because the data will not be released and strict confidentiality will be maintained.

(3) The research could not practicably be carried out without the waiver or alteration.

The research involves retrospective analysis of existing EHR data of a large number of patients. Without waiver, this analysis cannot be carried out.

(4) Whenever appropriate, the subjects will be provided with additional pertinent information after participation.

The subjects will be provided with relevant information on the analysis of the raw EHR data. When requested, publications resulting from the proposed study will also be shared with study participants.

## 3. RESULTS

### 3.1. Cohort Characteristics

We identified 20,699 patients with AKI at admission (i.e., within 48 hours of presentation) and without a prior or concurrent CKD diagnosis. Cohort details are outlined in Table 1, with cohort construction illustrated in Supplementary Figure S1. The median age was 63 years (IQR, 47-76), with 1,272 patients (6%) under 18; 10,617 (49%) were male. The time span from the start date of the initial AKI-associated hospitalization to the first diagnosis of CKD has a median of 152 days (IQR, 20-821.5), with mostly under 5 years of available data for 18,157 patients (88%). The first AKI visit years ranged from 1998 to 2023. Overall, 3,491 patients (17%) developed CKD, with significant age and self-reported race/ethnicity differences observed between CKD and non-CKD patients.

### 3.2. Patient Representation

We generated 967,034 patient vector representations (i.e., the concatenation of the temporal medical coding sequence embedding and the temporal creatinine embedding) across all patients in the cohort, each representing cumulative days within the entire study period for a patient. The model performance of the temporal medical coding sequence embedding is detailed in our previous work[37]. For temporal creatinine embeddings, our fine-tuned model achieved a test AUROC of 0.78 (95% CI: 0.75-0.80) and AUPRC 0.45 (95% CI: 0.40-0.50) for diagnosing CKD following AKI.

### 3.3. Clinical States from AKI to CKD and their Characteristics

We identified fifteen clinical states preceding terminal states. The number of states (clusters) was selected using the elbow method, with the elbow point occurring at 15 (See Supplementary Figure S2)[39]. This selection was further informed by expert review of the clinical characteristics of each cluster. Specifically, we engaged a nephrologist (JGN) to evaluate the interpretability and clinical plausibility of the resulting states. Clustering results were compared across several values near the elbow point, and the solution with 15 states was found to offer the most clinically meaningful, reasonable, and interpretable grouping. This combined analytical and expert-informed approach allowed us to balance model fit with clinical granularity, enabling the capture of nuanced patterns in disease progression. State labels are abbreviated (e.g., "State 1" is "S1"). To enhance understanding of the clustering results, we present illustrative examples from two representative patients in Figure 2. For each patient, we displayed their temporal medical code sequences, creatinine dynamics, the derived patient vectors, and the corresponding clinical states. These patient trajectories are highlighted on the Uniform Manifold Approximation and Projection

(UMAP)[61] visualization of patient vectors at key time points (start date and end date of the observation window, and state transitions). For example, one patient was initially assigned to S0 and transitioned to S12 at day 56 of follow-up. Another patient remained in S4 throughout the entire observation period. The complete UMAP visualization of the clustering results for all patient vectors in the cohort of AKI at admission is shown in Supplementary Figure S3.

Table 2 summarizes the predominant diseases associated with each state, grouped by organ system to highlight clinically meaningful patterns, together with their prevalence, demographic characteristics, and transition probabilities to CKD over time. These predominant diseases were derived from conditions with high state-specific prevalence and were curated and validated by a board-certified nephrologist (JGN) to ensure interpretability. Prevalence indicates the percentage of patient vectors assigned to each state with specific conditions. For example, S0 has a notably high prevalence of gastrointestinal and genitourinary diseases. Supplementary Table S7 presents a more detailed description of the state-specific prevalence of 481 clinically meaningful conditions such as ascites, bacteremia, acute on chronic systolic heart failure, respiratory failure, and cerebral infarction. For example, respiratory failure showed a prevalence of 34% in S0, 11% in S4, and 92% in S6. Supplementary Figure S4 presents the prevalence and hierarchy of conditions.

The majority of patients entered the cohort from initial state S0 (39%), S4 (30%), and S5 (29%) (see the summary statistics in Supplementary Table S8). Table 3 details the identified risk factors associated with CKD development for these initial states (the sensitivity analysis is shown in Supplementary Table S9). For example, in S0, baseline eGFR and the average serum sodium levels on day 1 were negatively associated with CKD.

### 3.4. Clinical State Transition Characterization

Figure 3A illustrates transition probabilities over a 5-year period. For instance, the transition probability from S4 to S5 (orange) after 1 year was 0.076 (95% CI: 0.071-0.082), increasing to 0.116 (0.109-0.124) after 5 years. The probability of remaining in S4 (light blue) decreased from 0.302 (0.289-0.314) after 1 year to 0.144 (0.133-0.154) after 5 years. Figure 3B and Table 2 provide transition probabilities to CKD development over time, with S3 having the highest (at year 1: 0.091 (0.075-0.107), at year 5: 0.294 (0.276-0.313), at year 10: 0.423 (0.4-0.445)).

A total of 15,607 patients (75%) experienced either no state transition (n=4,780, 23%) or only one transition (n=10,827, 52%) during the time window before reaching terminal states. Consequently, our analysis focused on transitions from the first to the second state. Initial state transitions with a prevalence greater than 5% are detailed in Supplementary Tables S8, S10, and S11, where we provide summary statistics of clinically meaningful variables and risk factors, thereby offering a clinically interpretable context for these transitions. Across these transitions, there was a consistent decrease in the probability of CKD development over time: from S0 to S12 (prevalence: 6%) at year 10, the probability decreased from 0.405 to 0.302; from S4 to S13 (6%), it decreased from 0.382 to 0.362; and from S5 to S10 (9%), it decreased from 0.411 to 0.34. Table 3 outlines the identified risk factors for CKD development for these initial state transitions (see Supplementary Table S9 for sensitivity analysis). For instance, in patients initially transitioning from S0 to S12, pre-existing DM was identified as a positive risk factor.

## 4. DISCUSSION

To characterize the progression from AKI to CKD, we analyzed the evolution of patients' health status over time. We identified fifteen distinct post-AKI clinical states, each with varying probabilities of CKD development. Overall, 17% of patients developed CKD. We also identified CKD risk factors across different AKI subpopulations based on their initial states or transitions between states.

### 4.1. Contributions

This study makes several key contributions. First, it provides a cost-effective approach to analyzing AKI-to-CKD progression by leveraging real-world longitudinal EHR data. Our method captures how medical conditions, treatments, and interventions evolve alongside creatinine dynamics, allowing for a more comprehensive understanding of CKD risk. This differs from the recently proposed hypergraph clustering approach for chronic disease trajectories in mild cognitive impairment[62], which focused exclusively on chronic conditions and did not incorporate non-chronic conditions, treatments, or biomarker dynamics. Our framework integrates the temporal sequences of all medical codes with disease-specific biomarker trajectories, providing a more comprehensive representation of patient health. Unlike prior work centered on chronic disease patterns, we explicitly model progression from an acute to a chronic state, exemplified by AKI-to-CKD transitions. Second, it applies multi-state modeling to EHR data to track patient transitions over time, providing a framework for estimating transition probabilities, identifying high-risk subgroups, and mapping common progression pathways. Third, we replicated known CKD risk factors from prior observational studies[10,63-68], reinforcing the validity of this cost-effective approach[69,70]. Finally, this study highlights how structured EHR data can be mined to extract clinically meaningful insights, offering an efficient and scalable method for identifying high-risk AKI patients who may benefit from early intervention and closer outpatient monitoring.

Our findings align with previous studies that have identified AKI severity, diabetes, hypertension, and congestive heart failure as key CKD risk factors.[10,63-65] Specifically, patients in **State 4 (S4)** with these conditions had an increased risk of CKD, as did patients in **S5** or those transitioning from **S4 to S13** with hypertension, and patients moving from **S0 to S12** with diabetes. Interestingly, among patients in **S0**, sepsis at admission appeared protective against CKD, consistent with Shum et al.,[66] who reported lower 90-day renal non-recovery in septic AKI patients. We also found that **S4 patients with liver disease or lower average magnesium levels** on day 1 had higher CKD risk, supporting Bassegoda et al.'s[67] findings on cirrhosis and AKI-to-CKD progression and Alves et al.'s[68] work linking hypomagnesemia to renal non-recovery.

In contrast to prior studies, we found that **younger patients in S4 with higher hemoglobin and albumin levels** on day 1 were more likely to develop CKD. This contradicts Chou et al.[71], who associated advanced age, lower hemoglobin, and lower albumin levels with CKD risk. Additionally, we uncovered novel CKD risk and protective factors across AKI subpopulations:

- In **S0**, higher **potassium** on day 1 increased CKD risk, while higher **sodium** was protective.
- In **S4**, higher **temperature and systolic blood pressure** on day 1 were risk factors, while higher **sodium and calcium levels** on day 1, and **RAS inhibitor use** within a year, were protective.
- In **S5**, higher **systolic blood pressure** on day 1 and **NSAID use** within a year were risk factors, while higher **diastolic blood pressure and sodium** were protective.
- For **patients transitioning from S0 to S12**, **NSAID use and higher body temperature** were linked to increased CKD risk.

These novel findings highlight potential AKI phenotypes that warrant further study.

### 4.2. Strengths and Limitations

This study has several strengths. By integrating longitudinal EHR data, it captures real-world patient trajectories and identifies high-risk AKI subpopulations in a cost-effective way. Additionally, the use of multi-state modeling allows for a more granular understanding of patient transitions compared to traditional observational research methods.

However, there are limitations. First, we did not incorporate urine-based AKI metrics due to their limited availability. Less than 10% of patients had at least two urine creatinine measurements within 48 hours of their first AKI visit. Second, EHR-based research is subject to data missingness and incompleteness, inconsistent clinical coding, and incomplete medication documentation – particularly for over-the-counter NSAIDs, which may affect our risk factor assessments. Relatedly, we defined the initial AKI visit as the first documented AKI-associated hospitalization in our system; however, we cannot rule out the possibility that patients may have had prior AKI episodes not captured in our data. Third, the median follow-up period was 152 days, with only 12% of patients having data beyond five years. While this exceeds the KDIGO-recommended three-month follow-up for AKI patients[72], longer observation may provide deeper insights. Fourth, while our patient representations were derived from embeddings, we undertook systematic steps to enhance interpretability by linking clinical states and transitions to clinically meaningful features. Despite this effort, we recognize that embedding-based methods are not fully transparent, and future research should continue to refine approaches that bridge advanced representation learning with clinical transparency. Additionally, this study was conducted at a single tertiary care center, which, despite its ancestrally and economically diverse patient population, may not be representative of broader healthcare settings.[73-80] Future work should validate these findings across multiple institutions to enhance generalizability. Finally, our study relies exclusively on structured EHR data, and unstructured clinical notes were not utilized. Incorporating such data in future work may further enrich patient representations and improve the characterization of disease progression.

### 4.3. Future Directions and Applications

This study demonstrates how a data-driven, scalable, and cost-effective method can characterize AKI subpopulations and identify CKD risk factors. With further validation, this framework can form the basis for automated decision support tools that integrate with EHR systems. These tools could help clinicians efficiently identify AKI patients at high risk for CKD and prioritize them for closer outpatient monitoring and early nephrology referral. As nephrology moves towards integrating multi-omics data, this methodology provides a foundation for extracting clinically meaningful insights from increasingly complex datasets, bridging the gap between big data analytics and real-time patient management.[81,82]

Beyond kidney disease, this framework has the potential to be adapted for other clinical conditions that involve progression from acute to chronic states. For instance, in the context of liver disease, it may be used to model the progression from acute liver injury or acute hepatitis to decompensated cirrhosis using biomarkers such as AST, ALT, INR, albumin, and platelet counts. This example underscores the versatility of our framework for studying disease progression in varied clinical contexts where both biomarker evolution and longitudinal clinical history play a central role. By combining temporal modeling of disease-specific biomarkers with representations of general clinical history derived from medical coding sequences, our approach offers a comprehensive view of patient trajectories.

## Data Availability

The clinical data used in this study contains Protected Health Information (PHI) and, as such, cannot be made readily available for distribution. Requests for access to the data will undergo review by the institutional IRB (Institutional Review Board) for consideration. The derived data underlying this article are available in the article and in its online supplementary material.

## Acknowledgments

The project was supported by grants from the National Library of Medicine (grant R01LM012895), the National Human Genome Research Institute (grants R01HG013031 and U01HG008680-09), the National Center for Advancing Translational Sciences (grants UL1TR001873-08, OT2TR003434, KL2TR001874 (JKH)), and the National Institute of Diabetes and Digestive and Kidney Diseases (grant K08DK132511 (JGN)). It was also supported by the American Society of Nephrology and Kidney Cure's Harold Amos Medical Faculty Development Award (JGN). The content is solely the responsibility of the authors and does not necessarily represent the official views of the National Institutes of Health.

## Author Contributions

The authors contributed to the study as follows: YF: Conceptualization, methodology, data curation, formal analysis, investigation, validation, visualization, and writing—original draft. JGN: Conceptualization, methodology, data curation, formal analysis, investigation, validation, and writing—original draft. CT: Methodology and writing—reviewing and editing. JKH: Data curation, validation, and writing—reviewing and editing. CW: Conceptualization, methodology, data curation, project administration, funding acquisition, resources, supervision, investigation, and writing—reviewing and editing.

**Competing Interests**

All authors declare that they have no known competing financial interests or personal relationships that could have appeared to influence the work reported in this paper.

## Tables

### Table 1. Characteristics of the AKI cohort at admission[a].

| Characteristics | Total patients (N=20,699) | Developed CKD (n=3,491) | Without CKD (n=17,208) | P value |
|---|---|---|---|---|
| Age at encounter, median (IQR), y | 63 (47-76) | 64 (52-75) | 63 (46-76) | <.001[c] |
| Sex | | | | |
| - Male | 10,617 (49) | 1,702 (49) | 8,375 (49) | |
| - Female | 10,077 (51) | 1,788 (51) | 8,829 (51) | 0.98[d,e] |
| - Unknown | 5 (0.0) | 1 (0.0) | 4 (0.0) | |
| Self-reported race | | | | |
| - White | 6,825 (33) | 1,179 (34) | 5,646 (33) | |
| - Black or African American | 3,481 (17) | 643 (18) | 2,838 (17) | |
| - Asian | 417 (2) | 66 (2) | 351 (2) | 0.01[d] |
| - Other[b] | 158 (0.8) | 20 (0.6) | 138 (0.8) | |
| - Unknown | 9,818 (47) | 1,583 (45) | 8,235 (48) | |
| Self-reported ethnicity | | | | |
| - Hispanic or Latino | 5,252 (25) | 1,045 (30) | 4,207 (24) | |
| - Not Hispanic or Latino | 8,118 (39) | 1,464 (42) | 6,654 (39) | <.001[d] |
| - Unknown | 7,329 (35) | 982 (28) | 6,347 (37) | |
| Time span from AKI to the first diagnosis of CKD, median (IQR), d | 152 (20-821.5) | 400 (100-1,231) | 111 (15-738.25) | <.001[c] |
| Start year of the visit, median (IQR), y | 2,017 (2,010-2,021) | 2,013 (2,008-2,018) | 2,018 (2,011-2,021) | <.001[c] |
| Patients requiring ICU admission during the visit | 3,219 (16) | 229 (7) | 2,990 (17) | <.001[d] |

[a] Data are presented as number (percentage) of patients unless otherwise indicated.

[b] The "Other" category of self-reported races includes "Native Hawaiian or Other Pacific Islander," "American Indian or Alaska Native," and "Other".

[c] Kolmogorov-Smirnov test.

[d] Chi-squared test.

[e] Statistical test was conducted on male and female categories, excluding unknown sex.

**Table 2. Characteristics of each clinical state: predominant diseases[a], demographics[b], and transition probabilities to CKD over time[c].**

| | Characteristics | Clinical State | | | | | | | | | | | | | | |
|---|---|---|---|---|---|---|---|---|---|---|---|---|---|---|---|---|
| | | S0 | S1 | S2 | S3 | S4 | S5 | S6 | S7 | S8 | S9 | S10 | S11 | S12 | S13 | S14 |
| **Predominant diseases** | **Systems** | | | | | | | | | | | | | | | |
| | Neurological | 4 | 5 | 2 | 4 | 2 | 3 | 3 | 4 | 4 | 4 | 3 | 1 | 3 | 3 | 3 |
| | Cardiovascular | 3 | 5 | 5 | 4 | 3 | 3 | 5 | 5 | 5 | 5 | 5 | 2 | 5 | 5 | 4 |
| | Respiratory | 3 | 5 | 5 | 5 | 3 | 3 | 5 | 5 | 5 | 5 | 5 | 2 | 5 | 5 | 5 |
| | Gastrointestinal | 5 | 5 | 5 | 5 | 3 | 5 | 5 | 5 | 5 | 5 | 4 | 2 | 5 | 5 | 5 |
| | Genitourinary | 5 | 4 | 3 | 5 | 3 | 4 | 5 | 5 | 4 | 5 | 3 | 2 | 5 | 3 | 3 |
| | Endocrine | 2 | 3 | 1 | 2 | 1 | 2 | 2 | 2 | 2 | 2 | 1 | 1 | 2 | 1 | 1 |
| | Immune | 2 | 4 | 2 | 3 | 1 | 1 | 3 | 3 | 3 | 3 | 2 | 1 | 3 | 2 | 2 |
| | Hematological | 2 | 3 | 2 | 3 | 1 | 1 | 3 | 3 | 3 | 3 | 2 | 1 | 3 | 2 | 2 |
| | Musculoskeletal | 2 | 3 | 1 | 2 | 1 | 2 | 2 | 3 | 2 | 2 | 1 | 1 | 2 | 2 | 1 |
| | Integumentary | 1 | 2 | 2 | 2 | 1 | 1 | 3 | 2 | 2 | 2 | 1 | 1 | 2 | 1 | 1 |
| | Metabolic | 3 | 4 | 3 | 3 | 2 | 3 | 5 | 5 | 5 | 5 | 3 | 1 | 5 | 3 | 2 |
| | Nutritional | 1 | 2 | 1 | 2 | 1 | 1 | 2 | 3 | 2 | 1 | 1 | 1 | 2 | 1 | 1 |
| | Congenital disorder | NA | 1 | 1 | 3 | 1 | 1 | 1 | 3 | 1 | 1 | 1 | 1 | 1 | 1 | 1 |
| | No. (%) of patient vector representations | 182,725 (19) | 16,338 (2) | 44,793 (5) | 50,046 (5) | 116,359 (12) | 186,961 (19) | 26,599 (3) | 16,365 (2) | 40,250 (4) | 47,838 (5) | 80,702 (8) | 28,255 (3) | 55,279 (6) | 49,212 (5) | 25,312 (3) |
| | No. (%) of patients | 8,258 (40) | 349 (2) | 667 (3) | 765 (4) | 6,195 (30) | 7,410 (36) | 213 (1) | 374 (2) | 1,159 (6) | 2,233 (11) | 3,347 (16) | 919 (4) | 1,440 (7) | 1,409 (7) | 660 (3) |
| **Demographics** | Age, median (IQR), y | 63.0 (50.0-74.0) | 39.0 (21.0-55.0) | 0.0 (0.0-4.0) | 50.0 (27.0-64.0) | 62.0 (51.0-75.0) | 62.0 (50.0-73.0) | 2.0 (0.0-6.0) | 68.0 (58.0-77.0) | 58.0 (40.0-72.0) | 60.0 (52.0-70.0) | 58.0 (45.0-70.0) | 56.0 (35.0-71.0) | 57.0 (41.0-68.0) | 56.0 (41.0-69.0) | 61.0 (47.0-73.0) |
| | Sex (Male) | 4,075 (49) | 215 (62) | 363 (54) | 325 (42) | 2,987 (48) | 3,536 (48) | 117 (55) | 147 (39) | 427 (37) | 1,338 (60) | 1,773 (53) | 483 (53) | 686 (48) | 702 (50) | 308 (47) |
| | **Self-reported race** | | | | | | | | | | | | | | | |
| | White | 3,200 (39) | 130 (37) | 240 (36) | 307 (40) | 1,580 (26) | 2,431 (33) | 74 (35) | 144 (39) | 451 (39) | 809 (36) | 1,252 (37) | 162 (18) | 535 (37) | 403 (29) | 183 (28) |
| | Black or African American | 1,582 (19) | 59 (17) | 118 (18) | 145 (19) | 796 (13) | 1,319 (18) | 44 (21) | 41 (11) | 192 (17) | 426 (19) | 483 (14) | 57 (6) | 215 (15) | 187 (13) | 67 (10) |
| | Asian | 225 (3) | 11 (3) | 39 (6) | 31 (4) | 71 (1) | 131 (2) | 19 (9) | 12 (3) | 34 (3) | 49 (2) | 78 (2) | 4 (0.4) | 62 (4) | 22 (2) | 3 (0.5) |
| | Other | 24 (0.3) | 4 (1) | 12 (2) | 6 (0.8) | 89 (1) | 67 (0.9) | 1 (0.5) | 3 (0.8) | 11 (0.9) | 12 (0.5) | 31 (0.9) | 16 (2) | 2 (0.1) | 28 (2) | 18 (3) |

| Characteristics | | S0 | S1 | S2 | S3 | S4 | S5 | S6 | S7 | S8 | S9 | S10 | S11 | S12 | S13 | S14 |
|---|---|---|---|---|---|---|---|---|---|---|---|---|---|---|---|---|
| | | | | | | | | | | | | | | Clinical State | | |
| | Unknown | 3,227 (39) | 145 (42) | 258 (39) | 276 (36) | 3,659 (59) | 3,462 (47) | 75 (35) | 174 (47) | 471 (41) | 937 (42) | 1,503 (45) | 680 (74) | 626 (43) | 769 (55) | 389 (59) |
| Self-reported ethnicity | Hispanic or Latino | 2,418 (29) | 68 (19) | 180 (27) | 197 (26) | 1,074 (17) | 2,193 (30) | 62 (29) | 96 (26) | 301 (26) | 645 (29) | 705 (21) | 41 (4) | 450 (31) | 221 (16) | 94 (14) |
| | Not Hispanic or Latino | 4,549 (55) | 170 (49) | 357 (54) | 369 (48) | 1,363 (22) | 2,680 (36) | 116 (54) | 158 (42) | 475 (41) | 1,030 (46) | 1,297 (39) | 113 (12) | 753 (52) | 371 (26) | 119 (18) |
| | Unknown | 1,291 (16) | 111 (32) | 130 (19) | 199 (26) | 3,758 (61) | 2,537 (34) | 35 (16) | 120 (32) | 383 (33) | 558 (25) | 1,345 (40) | 765 (83) | 237 (16) | 817 (58) | 447 (68) |
| Transition probability to CKD at time t (95% CI) | 6 months | 0.074 (0.068-0.08) | 0.045 (0.013-0.077) | 0.053 (0.042-0.063) | 0.091 (0.075-0.107) | 0.085 (0.078-0.093) | 0.074 (0.068-0.08) | 0.039 (0.012-0.066) | 0.058 (0.031-0.084) | 0.049 (0.03-0.068) | 0.072 (0.065-0.079) | 0.07 (0.063-0.077) | 0.061 (0.05-0.071) | 0.048 (0.04-0.057) | 0.094 (0.08-0.108) | 0.057 (0.034-0.081) |
| | 1 year | 0.109 (0.101-0.116) | 0.067 (0.025-0.108) | 0.083 (0.07-0.096) | 0.135 (0.118-0.151) | 0.125 (0.116-0.133) | 0.113 (0.105-0.12) | 0.064 (0.031-0.096) | 0.085 (0.054-0.116) | 0.076 (0.048-0.105) | 0.108 (0.098-0.119) | 0.104 (0.095-0.112) | 0.089 (0.076-0.102) | 0.077 (0.067-0.087) | 0.134 (0.117-0.151) | 0.084 (0.05-0.118) |
| | 3 years | 0.208 (0.194-0.222) | 0.113 (0.05-0.175) | 0.161 (0.142-0.181) | 0.228 (0.21-0.245) | 0.212 (0.201-0.222) | 0.205 (0.195-0.215) | 0.121 (0.073-0.168) | 0.14 (0.099-0.18) | 0.144 (0.094-0.194) | 0.196 (0.175-0.216) | 0.18 (0.17-0.19) | 0.144 (0.127-0.162) | 0.154 (0.138-0.17) | 0.219 (0.197-0.241) | 0.139 (0.083-0.195) |
| | 5 years | 0.282 (0.26-0.305) | 0.139 (0.061-0.217) | 0.214 (0.188-0.24) | 0.294 (0.276-0.313) | 0.274 (0.262-0.285) | 0.274 (0.263-0.286) | 0.152 (0.089-0.215) | 0.185 (0.135-0.235) | 0.19 (0.124-0.257) | 0.26 (0.23-0.29) | 0.235 (0.223-0.246) | 0.189 (0.169-0.209) | 0.206 (0.185-0.226) | 0.272 (0.246-0.298) | 0.178 (0.106-0.25) |
| | 10 years | 0.405 (0.382-0.427) | 0.19 (0.085-0.296) | 0.313 (0.279-0.348) | 0.423 (0.4-0.445) | 0.382 (0.368-0.396) | 0.411 (0.395-0.427) | 0.214 (0.121-0.307) | 0.267 (0.2-0.334) | 0.281 (0.183-0.38) | 0.385 (0.331-0.438) | 0.34 (0.325-0.355) | 0.261 (0.235-0.286) | 0.302 (0.276-0.328) | 0.362 (0.329-0.394) | 0.244 (0.145-0.342) |

[a] Predominant disease characteristics by patient clinical states. Diseases were categorized by organ systems, and the category of congenital disorders was included as well. Disease prevalence rate within a clinical state is indicated by labels, 1: <50%, 2: 50-74%; 3: 75-89%; 4: 90-94%; 5: ≥95%. For example, clinical state S1 has label 5 for neurological diseases, indicating that >=95% of the vectors assigned to state S1 represent a specific time interval during which the related condition codes were recorded.

[b] Demographic characteristics by patient clinical states. Age (median and IQR) was calculated based on the ages at the end date of the intervals of patient vectors that were assigned to a specific state. Each interval extended one day longer than the previous, starting from the start date of the first AKI visit of a patient and spanning within the entire time window. Statistics of sex and self-reported race and ethnicity were calculated for unique patients who had vectors that were assigned to a specific state at least once during the entire time window.

[c] Transition probabilities from a clinical state to CKD over time were estimated using the Aalen-Johansen estimator.

**Table 3. CKD risk factors for each prevalent initial clinical state and initial state transition, identified using weighted Cox proportional hazards models.**

| Initial state/Initial state transition | Variable | | All patients with identical initial state/initial state transition | | | N |
|---|---|---|---|---|---|---|
| | Descriptor | Phenotypic traits | Average hazard ratio | Uncorrected p-value | Corrected p-value | |
| S0 | Baseline | eGFR (mL/min/1.73 m²) | 0.99 (0.98-0.99) | <.001 | 0.02 | 8,037 |
| | Avg. value at Day 1 | Na (mEq/L) | 0.96 (0.93-0.99) | 0.004 | 0.04 | |
| | | K (mEq/L) | 1.35 (1.10-1.66) | 0.005 | 0.04 | |
| | Day 1 | Sepsis | 0.49 (0.30-0.79) | 0.004 | 0.04 | |
| S4 | Pre-existing | CHF | 1.55 (1.32-1.83) | <.001 | <.001 | 6,123 |
| | | DM | 1.55 (1.29-1.85) | <.001 | <.001 | |
| | | HTN | 1.51 (1.30-1.76) | <.001 | <.001 | |
| | | LD | 1.43 (1.20-1.70) | <.001 | <.001 | |
| | | RAS inhibitors | 0.77 (0.65-0.92) | 0.003 | 0.007 | |
| | Baseline | eGFR (mL/min/1.73 m²) | 0.99 (0.99-0.99) | <.001 | <.001 | |
| | At encounter | Age (y) | 0.99 (0.99-1.00) | <.001 | 0.002 | |
| | Avg. value at Day 1 | Temp. (°C) | 1.78 (1.44-2.20) | <.001 | <.001 | |
| | | SBP (mmHg) | 1.02 (1.01-1.02) | <.001 | <.001 | |
| | | Na (mEq/L) | 0.98 (0.97-0.99) | <.001 | <.001 | |
| | | Ca (mg/dL) | 0.91 (0.84-0.98) | 0.01 | 0.03 | |
| | | Mg (mg/dL) | 0.72 (0.59-0.89) | 0.002 | 0.005 | |
| | | Alb. (g/dL) | 1.26 (1.11-1.44) | <.001 | 0.002 | |
| | | ALP (IU/L) | 1.00 (1.00-1.00) | 0.002 | 0.005 | |
| | | Hb (g/dL) | 1.05 (1.02-1.09) | 0.002 | 0.005 | |
| S5 | Pre-existing | HTN | 2.47 (1.83-3.32) | <.001 | <.001 | 5,907 |
| | | NSAIDs | 1.47 (1.13-1.90) | 0.004 | 0.02 | |
| | Baseline | eGFR (mL/min/1.73 m²) | 0.99 (0.98-1.00) | <.001 | 0.004 | |
| | Avg. value at Day 1 | SBP (mmHg) | 1.02 (1.01-1.02) | <.001 | <.001 | |
| | | DBP (mmHg) | 0.98 (0.97-0.99) | 0.007 | 0.04 | |
| | | Na (mEq/L) | 0.97 (0.95-0.99) | <.001 | 0.004 | |
| S0->S12 | Pre-existing | DM | 2.56 (1.49-4.41) | <.001 | 0.02 | 1,271 |
| | | NSAIDs | 2.84 (1.51-5.34) | 0.001 | 0.02 | |
| | Avg. value at Day 1 | Temp. (°C) | 2.50 (1.39-4.52) | 0.002 | 0.02 | |

| Initial state/Initial state transition | Variable | | All patients with identical initial state/initial state transition | | | |
|---|---|---|---|---|---|---|
| | Descriptor | Phenotypic traits | Average hazard ratio | Uncorrected p-value | Corrected p-value | N |
| S4->S13 | Day 1 | Sepsis | 0.36 (0.20-0.68) | 0.001 | 0.02 | 1,317 |
| | Pre-existing | CHF | 1.68 (1.18-2.40) | 0.004 | 0.04 | |
| | At encounter | Age (y) | 0.99 (0.98-1.00) | 0.001 | 0.02 | |
| | Avg. value at Day 1 | SBP (mmHg) | 1.04 (1.02-1.06) | <.001 | 0.003 | |

Abbreviations: CKD: chronic kidney disease; CHF: congestive heart failure; DM: diabetes mellitus; HTN: hypertension; LD: liver disease; SBP: systolic blood pressure; DBP: diastolic blood pressure; Na: sodium; K: potassium; Ca: calcium; Mg: magnesium; Alb.: albumin; ALP: alkaline phosphatase; Hb: hemoglobin; eGFR: estimated glomerular filtration rate; Day 1: the start date of the first AKI visit; N: the number of patients; CI: confidence interval.

**Figure Legends**

**Figure 1. Procedure for characterizing AKI-to-CKD progression.**

**Figure Legend:** Medical codes (e.g., conditions, drugs, procedures) were organized chronologically and transformed into sequences. These sequences were embedded to derive vectors. Creatinine (Cr) time series were also embedded and concatenated (combined) with the sequence vectors to form a comprehensive patient vector representing health status over time (i.e., cumulative days within the entire time window). The time window extends from the start of the first AKI hospitalization to either the earliest CKD diagnosis or the last recorded database entry for patients with right-censored data or competing risks of death. Clinical states, reflecting patient health status, were extracted through clustering, and transitions between states or to CKD development were identified.

**Figure 2. Example patient trajectories showing temporal medical codes, creatinine dynamics, clinical states, and transitions.**

**Figure Legend:** Illustrative examples from two representative patients are shown. For each patient, temporal medical code sequences and creatinine trajectories were embedded into patient vectors and linked to clinical states. Post-AKI trajectories are highlighted on the Uniform Manifold Approximation and Projection (UMAP) visualization of the cohort at selected time points (start date and end date of the observation window, and state transitions). For clarity, only the UMAP entries corresponding to the states that each patient was assigned to are displayed. For example, for the first patient, only vectors from States 0 and 12 across the entire cohort and time horizon are shown.

**Figure 3. Transition probabilities between clinical states over 0 to 5 years.**

**Figure Legend:** Two terminal states - CKD development and the competing risk of death - are included. (A) Transition probabilities between any two states over 0 to 5 years. For example, the upper right subfigure, shows the transition probability from State 4 (S4) to State 5 (S5; orange) was 0.076 at year 1 and increased to 0.116 5 years after entering the cohort. The same subfigure shows the probability of remaining in State 4 (S4; light blue) was 0.302 at year 1 and decreased to 0.144 after 5 years. (B) Transition probabilities from a state to a terminal state over time, with 95% confidence intervals (CIs) shown by grey lines.

# Figures

## Figure 1. Procedure for characterizing AKI-to-CKD progression.

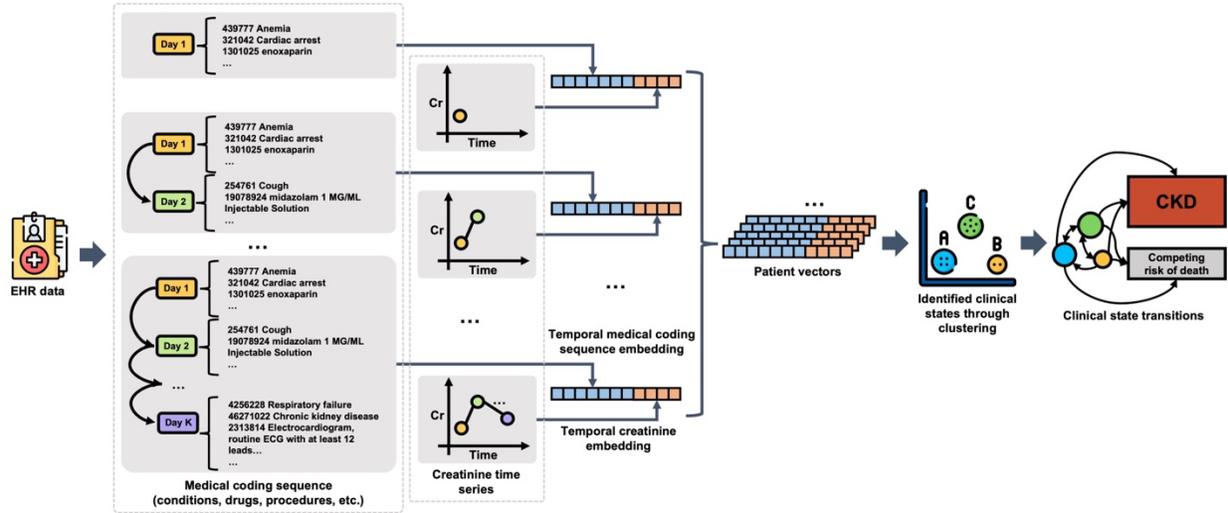

**Figure 2. Example patient trajectories showing temporal medical codes, creatinine dynamics, clinical states, and transitions.**

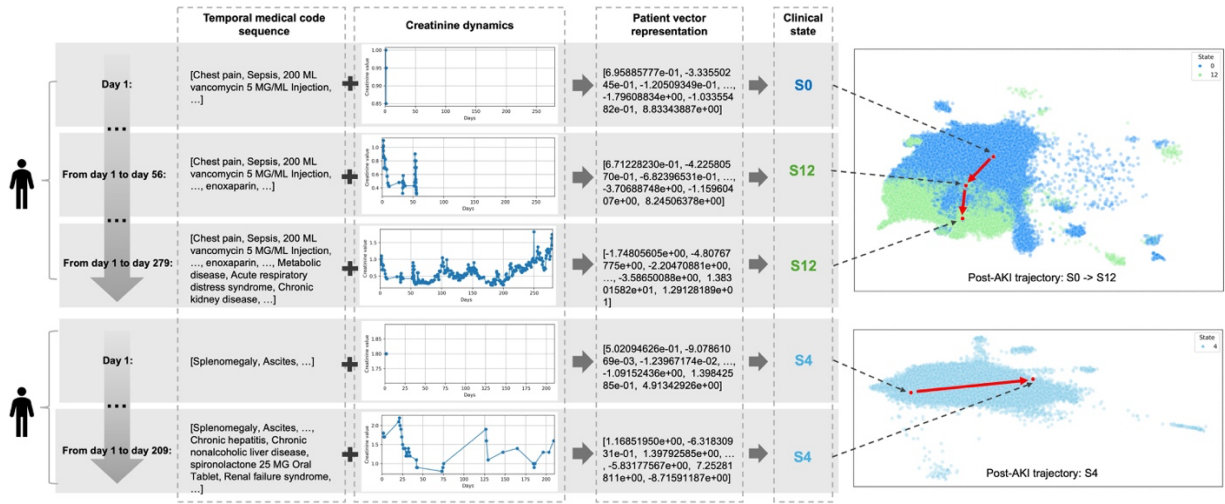

**Figure 3. Transition probabilities between clinical states over 0 to 5 years.**

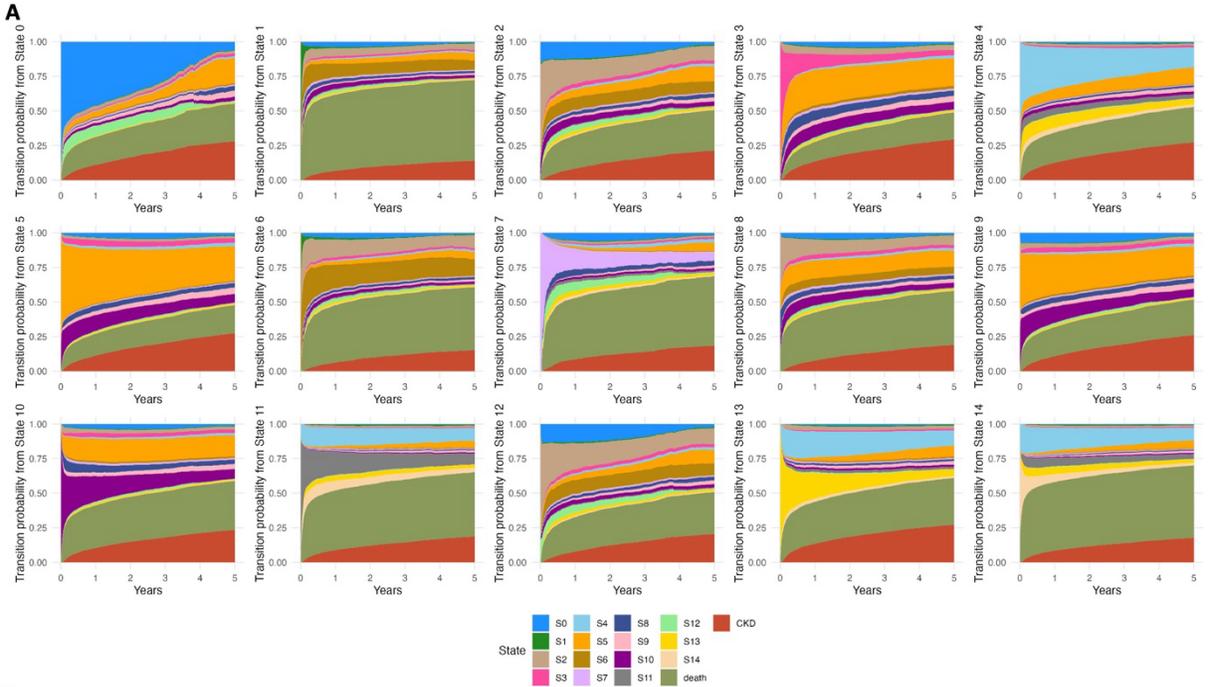
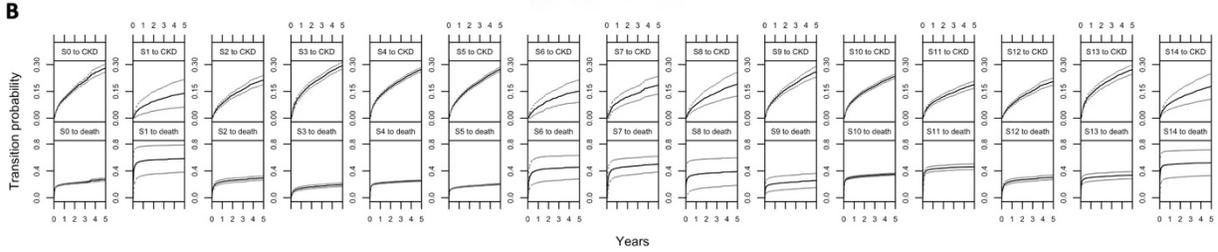